\documentclass[letterpaper, 10 pt, conference]{ieeeconf}  
\IEEEoverridecommandlockouts                              

\usepackage{graphics} 
\usepackage{epsfig} 
\usepackage{mathptmx} 
\usepackage{times} 
\usepackage{amsmath} 
\usepackage{amssymb}  
\usepackage{subcaption}
\usepackage{bm}
\usepackage{hyperref}

\usepackage{algorithm, algorithmicx}
\usepackage[noend]{algpseudocode}

\newcommand{\Ls}{\eta}

\title{\LARGE \bf
Automated Testing with Temporal Logic Specifications for Robotic Controllers using Adaptive Experiment Design*
}

\author{Craig Innes$^{1}$ and Subramanian Ramamoorthy${^1}$
\thanks{$^{1}$ At the Institute of Perception, Action and Behaviour (IPAB),
        University of Edinburgh, United Kingdom. Email:
        {\tt\small craig.innes@ed.ac.uk, s.ramamoorthy@ed.ac.uk}}%
\thanks{*Work supported by a grant from the UKRI Strategic Priorities Fund to the UKRI Research Node on Trustworthy Autonomous Systems Governance and Regulation (EP/V026607/1, 2020-2024)
}}

\begin{document}

\maketitle
\thispagestyle{empty}
\pagestyle{empty}

\begin{abstract}

Many robot control scenarios involve assessing system robustness against a task specification. If either the controller or environment are composed of ``black-box'' components with unknown dynamics, we cannot rely on formal verification to assess our system. Assessing robustness via exhaustive testing is also often infeasible if the number of possible environments is large compared to experiment cost. 

Given limited budget, we provide a method to choose experiment inputs which accurately reflect how robustly a system satisfies a given specification across the domain. By combining signal temporal logic metrics with adaptive experiment design, our method chooses each experiment by incrementally constructing a surrogate model of the specification robustness. This model then chooses experiments in areas of either high prediction error or high uncertainty.

Our evaluation shows how this adaptive experiment design results in sample-efficient descriptions of system robustness. Further, we show how to use the constructed surrogate model to assess the behaviour of a data-driven control system under domain shift.

\end{abstract}

\section{Introduction}
\label{sec:introduction}

Suppose you must design a controller for a robotic manipulator which moves objects between containers in a warehouse. Such scenarios often consist of two parts:

First is a  \emph{task specification} $\varphi$ (e.g., ``\emph{Move all the objects from the `in' container to the `out' container within one minute}'' or ``\emph{Never drop fragile objects on the floor}''). Most common robot task specifications can be written formally in \emph{signal temporal logic} (\textsc{stl}) \cite{maler2004monitoring, menghi2019specification}. Such logics also come with a \emph{robustness metric} --- a function which takes the output trajectory of a system and returns a numeric value representing how robustly $\varphi$ was satisfied (large positive values correspond to robust satisfaction; negative to failure).

Second is a \emph{domain specification} $\mathcal{D}$ --- the range of environments our controller could operate in (e.g., ``\emph{Containers are within 1 metre of the robot}'', ``\emph{Objects weigh 50-100g}'').

We would like to ensure our controller robustly satisfies $\varphi$ across all environment inputs $x \in \mathcal{D}$. Ideally, we could use rigorous formal methods to prove our controller satisfies $\varphi$. Unfortunately, though it is usually possible to write $\varphi$ formally, such rigorous verification methods typically also require a well-defined, tractable mathematical model of our \emph{controller} and \emph{environment}. In reality, both controller and environment may have intractable dynamics, or contain data-driven ``black-box'' components whose mathematical formulations are completely unknown. 

Absent a formal proof, we could ensure our controller satisfies $\varphi$ by testing it against every $x \in \mathcal{D}$. Again, unfortunately, this is typically impossible: domains with more than a few parameters require a combinatorial number of tests (continuous parameters like object position have a technically infinite number of values). Further, simulating even a single instance of the domain may take non-trivial time; a real-world test may take even longer. Inevitably, we only have time to test our controller in a finite subset of $\mathcal{D}$.

Given a budget of $N$ experiments, how can we choose the best subset of environment inputs to get a representative view of how robustly our controller satisfies $\varphi$ across $\mathcal{D}$? We do not just mean \emph{falsification} (finding a $x \in \mathcal{D}$ where we fail to satisfy $\varphi$). We may also wish to analyse where our system is \emph{close} to failing, or areas where $\varphi$ is consistently satisfied.

One simple testing strategy would be to do $N$ experiments at random. It is unlikely such a sample would offer good coverage over the range of parameters in $\mathcal{D}$, nor their interactions---random samples tend to have high \emph{discrepancy}. Alternatively, we could spread our $N$ experiments evenly across $\mathcal{D}$ by generating a low-discrepancy \emph{uniform design} \cite{garud2017design}. This is a vastly superior method for achieving good coverage of $\mathcal{D}$ (and in previous work, we show how to synthesize such designs automatically from a declarative specification of robot environments \cite{innes2021automatic}). However, such uniform designs fail to leverage any information about $\varphi$---for large areas of $\mathcal{D}$, the robustness of $\varphi$ may be identical; in smaller areas, it may vary wildly. As a result, uniform designs often ``waste" experiments by focusing on large ``uninteresting'' areas of $\mathcal{D}$.

This paper's key idea is to frame the choosing of $N$ experiments to test black-box controller robustness as the task of \emph{learning a model} $\hat{f}$ from environment parameters in $\mathcal{D}$ to a robustness metric built around $\varphi$.

Our algorithm combines quantitative \textsc{stl} metrics (\ref{sec:stl}), gaussian processes (\ref{sec:surrogate-model}), and adaptive experiment design (\ref{sec:adaptive-experiment-design}) to construct an online model $\hat{f}$. In this way, we can choose successive experiments which construct a sample-efficient robustness across $\mathcal{D}$ based on where $\hat{f}$ currently believes is the ``most interesting" part of the space.

In experiments on simulated reaching, pick-and-place, and sliding tasks on a 7-DOF robot arm (\ref{sec:experiments}), we show how our method efficiently tests satisfaction of $\varphi$ across $\mathcal{D}$ by building accurate models of task robustness. We find that our method consistently outperforms models built using random experiments or uniform designs. By applying our testing method on a reinforcement learning controller from the OpenAIGym Robotics suite \cite{brockman2016openai}, we also show how our algorithm can be utilized to analyse trouble spots which arise under domain shift (\ref{sec:experiments:slide}).

\section{Learning Surrogate Models of \textsc{stl} Specifications on a Budget}
\label{sec:method}

Let us outline the main components of our task and our aim: Our \emph{domain} is a $d$-dimensional space $\mathcal{D} \subseteq \mathbb{R}^d$. Our expensive black-box model of our controller and environment is $g$. For a given $x \in \mathcal{D}$, $g(x)$ outputs a trajectory $\tau = [s_1, \dots, s_T]$, where $s_{i}$ is the system state at time $i$.

We also have a task specification $\varphi$---an \textsc{stl} formula---and robustness metric $\Ls$. Running $\tau$ through $\Ls$ gives us $y = \Ls(\varphi, \tau)$---a robustness score. Positive scores mean $\varphi$ was satisfied, negative scores that it was not. We wish to understand the relationship between inputs $x \in \mathcal{D}$ and satisfaction of our specification $\varphi$. That is, we want to learn the function $f(x) = \Ls(\varphi, g(x))$. Since we do not know the true form of $f$, we wish to estimate its behaviour by learning a function $\hat{f}$ using input-output pairs $X = \{ x^{(1)} , \dots \ x^{(N)}\}$, $\bm{y} = \{y^{(1)}, \dots, y^{(N)}\}$, where $N$ is our experiment budget.

Our goal is to choose $X \subset \mathcal{D}$ such that $\hat{f}$ accurately describes the behaviour of $f$ across $\mathcal{D}$. We choose each experiment \emph{sequentially}, giving us an opportunity to leverage the predictions of $\hat{f}$ to inform each choice.

We first summarize how to learn $\hat{f}$ using Gaussian Processes (\ref{sec:surrogate-model}). We then describe our contribution, which involves formulating our task specification $\varphi$ in \textsc{stl} with an associated quantitative robustness function $\Ls$ (\ref{sec:stl}). We then use $\varphi$ and $\Ls$ to drive an adaptive sequential experiment design, leveraging the bias and variance from from $\hat{f}$ to select informative next experiments (\ref{sec:adaptive-experiment-design}).

\subsection{Surrogate Modelling}
\label{sec:surrogate-model}

We wish to estimate the underlying function $f : \mathcal{D} \rightarrow \mathbb{R}$. Let us assume $f$ was drawn from a space of possible functions generated via a Gaussian Process (\textsc{gp}) prior: 

\begin{gather}
    f(\cdot) \sim GP(0, k_{\theta}(\cdot , \cdot))
    \label{eqn:gp}
\end{gather}

Here, $k_{\theta}(\cdot, \cdot)$ is a \emph{kernel} function parameterized by $\theta$. For example, the Squared Exponential (\textsc{se}) kernel has the form:

\begin{equation}
    k_{\{\sigma, \ell\}}(x, x') = \sigma^2\exp\left(-\frac{(x - x')^2}{2\ell}\right)
    \label{eqn:squared-exp}
\end{equation}

Given data $\langle X, \bm{y} \rangle$ we can learn the best-fitting $\hat{f}(\cdot)$ by choosing values of $\theta$ which maximize the data likelihood:

\begin{gather}
    \hat{f}(\cdot) = \underset{\theta}{\mathrm{argmax}}\ p(\bm{y} | X, \theta) \\
    p(\bm{y} | X, \theta) = \mathcal{N}(\bm{y}; 0, K_{\theta}(X, X))
    \label{eqn:fit-gp}
\end{gather}

Where $K_{\theta}(X,X)_{ij} = k_{\theta}(x^{(i)}, x^{(j)})$. We then predict value $y^{(*)}$ of a new point $x^{(*)}$ and its variance $s^2(x^{(*)})$ using standard results \cite{williams2006gaussian}, with $K = K_{\theta}(X, X)$, $\bm{k}^{(*)}_i = k_{\theta}(x^{(*)}, x^{(i)})$:

\begin{gather}
    y^{(*)} = \bm{k}_{\theta}^{(*)\intercal} K^{-1} \bm{y} \label{eqn:gp-mean} \\
    s^2(x^{(*)}) = k_{\theta}(x^{(*)}, x^{(*)}) - \bm{k}^{(*)\intercal}K^{-1}\bm{k}^{(*)} \label{eqn:gp-var}
\end{gather}

\subsection{\textsc{stl} and robustness metrics}
\label{sec:stl}

In robotics, task specifications involve properties on continuous signals over time. Signal temporal logic (\textsc{stl}) is a standard formal language for expressing such properties. \textsc{stl} is defined by the following grammar:

\begin{equation}
    \begin{aligned}
        &\varphi := \top \mid \pi \mid \neg \varphi \mid \varphi_1 \wedge \varphi_2 \mid \square_{I} \varphi \mid \diamondsuit_{I} \varphi \mid \varphi_1 \mathcal{U}_{I} \varphi_2 \\
    \end{aligned}
\label{eqn:stl-syntax}
\end{equation}

Where $\pi$ is a predicate which can be expressed in the form $h(x) - u \leq 0$ where $h: \mathbb{R}^n \rightarrow [-1, 1]$ and $u \in [-1, 1]$. Symbol $\square_{I} \varphi$ means  $\varphi$ is \emph{always} true at every time-step within interval $I$. $\diamond_{I} \varphi$  means $\varphi$ must \emph{eventually} be true at some time-step in $I$. $\varphi_{1} \mathcal{U}_{I} \varphi_{2}$ means $\varphi_1$ must remain true within $I$ \emph{until} $\varphi_{2}$ becomes true. As an example, for the \textsc{stl} formula:

\begin{equation}
    \diamondsuit_{[0, 5]} \left( r_{x} \geq 3.0 \right)
    \label{eqn:stl-example}
\end{equation}

$r_{x}$ must achieve a value of at least 3.0 within 5 seconds. Given an \textsc{stl} specification $\varphi$ and trajectory $\tau$, we can now trivially check whether $\tau$ satisfies or violates $\varphi$. However, for building robust systems, it is often far more useful to know not just binary satisfaction/violation, but \emph{how well} (or badly) we satisfied $\varphi$. Ideally, such a metric $\Ls$ should capture both \emph{temporal} and \emph{spatial} robustness. To illustrate, consider that in (\ref{eqn:stl-example}), if $r_{x}$ took 4.9 seconds to go above 3, we would consider this less robust than if it took 1 second. Similarly, if $r_{x}$ only ever achieved a value of 3.1, we would consider that less robust than if it reached 10. For learning a model $\hat{f}$, we seek two further characteristics. First, $\Ls$ should be \emph{sound}---Positive values always imply $\tau$ satisfies the original $\varphi$ formula; negative values always imply $\tau$ violates them. Second, $\Ls$ should be \emph{smooth}---if $\tau$ and $\tau'$ are similar, $\Ls(\varphi, \tau)$ and $\Ls(\varphi, \tau')$ should be similar. For these reasons, we set our metric $\Ls$ to be the Arithmetic-Geometric Mean robustness (\textsc{agm}) \cite{mehdipour2019arithmetic}. We present a subset of the cases of $\Ls$ below:

\begin{align}
        &\Ls(\top, \tau, t) = 1 \\
        &\Ls(\pi,\tau,t) = \frac{1}{2} (h(\tau[t]) - u) \\
        &\Ls(\neg \varphi, \tau, t) = -\Ls(\varphi, \tau, t) \\
        &\Ls(\bigwedge_{i = 1}^m \varphi_i , \tau, t) = \begin{cases}
            \sqrt[m]{\prod_i 1 + \Ls^{\varphi_i}_{t} } - 1 \enskip &\forall i, \Ls^{\varphi_i}_{t} > 0\\
            \frac{1}{m}\sum_{i} [\Ls^{\varphi_i}_{t}]_{-} \enskip &otherwise
        \end{cases} \\
        &\Ls(\square_{I} \varphi, \tau, t) = \begin{cases}
            \sqrt[N]{\prod\limits_{k \in I} 1 + \Ls^{\varphi}_{t+k}} - 1 \enskip &\forall k . \Ls^{\varphi}_{t+k} > 0\\
            \frac{1}{N} \sum_{k \in I} [\Ls^{\varphi}_{t+k}]_{-} \enskip & otherwise
        \end{cases}
\end{align}

Where $\eta^{\varphi}_{t}$ is an abbreviation of $\eta(\varphi, \tau, t)$, $[f]_{+} = max(0, f)$, and $[f]_{-} = -[-f]_{+}$. Formulas $\Ls(\diamond_{I} \varphi)$ and $\Ls(\varphi_1 \mathcal{U}_{I} \varphi_2)$ follow similar patterns---geometric means when all sub-formulae are satisfied, arithmetic means otherwise. For details, see \cite{mehdipour2019arithmetic}. 

\subsection{Adaptive Experiment Design}
\label{sec:adaptive-experiment-design}

We now have a metric $\Ls$ for \textsc{stl} specification $\varphi$, and a method to construct $\hat{f}$ from data $\langle X, \bm{y} \rangle$. How should we choose experiments $1 \dots N$ to construct an accurate model of $f$? Candidates from $\mathcal{D}$ most likely to be useful in improving $\hat{f}$ are those exhibiting high \emph{bias} (points where error $(f(x) - \hat{f}(x))^2$ is large), and those which exhibit high \emph{variance} (points where $s^2(x)$ is large) \cite{fuhg2021state}.

We get the variance term $s^2(x)$ directly via (\ref{eqn:gp-var}), but not the bias---calculating the true error for every candidate $x$ would require evaluating $f(x)$. This is precisely the expensive operation we are attempting to avoid. However, we can approximate this error using a form of leave-one-out cross validation (\textsc{cv}). At the current time step, we can use any of the previous experiments $x^{(i)}$ as a validation point, calculating the \textsc{CV}-error as:

\begin{equation}
    e^2_{CV}(x^{(i)}) = (y^{(i)} - \hat{f}^{-i}(x^{(i)}))^2
    \label{eqn:ecv}
\end{equation}

Where $\hat{f}^{-i}$ is a model fit using all points currently in $X$ \emph{except} $x^{(i)}$. In words, this calculates our error in predicting $f(x^{(i)})$ if $x^{(i)}$ was absent from our data. To predict the error of an \emph{arbitrary point} $x$, we make a strong assumption --- nearby points will have similar error. Formally:

\begin{equation}
    e^2_{CV}(x) \approx e^2_{CV}\left( \underset{x^{*} \in [x^{(1)}\dots x^{(i})}{\mathrm{argmin}} \lVert x - x^* \rVert \right).
\end{equation}

Combining $e^2_{CV}(x)$ and $s^2(x)$, we arrive at a metric from kriging---the Expected Prediction Error (\textsc{epe}) \cite{liu2017adaptive}:

\begin{equation}
    \textsc{epe}^{\alpha}(x) = \alpha \times e^2_{CV}(x) + (1 - \alpha) \times s^2(x)
    \label{eqn:epe}
\end{equation}

Where $\alpha$ balances locally exploiting areas of high expected error with globally exploring areas of high model variance. At each step $t$, we set $\alpha$ as:

\begin{equation}
    \alpha = 0.99 \times min [ 1, \ 0.5 \times \frac{y^{(t-1)} - \hat{f}(x^{(t - 1)})}{e^2_{CV}(x^{(t - 1)})}]
    \label{eqn:mepe-alpha}
\end{equation}

Equation (\ref{eqn:mepe-alpha}) favours global exploration if the \textsc{cv}-error over-estimated the true error at $x^{(t-1)}$, and local exploitation if it under-estimated the true error.

\subsection{Choosing an initial set of starting experiments}
\label{sec:starting-exps}

\newcommand{\uniD}{\mathcal{U}^{N}_d}

One issue with selecting new experiments to improve $\hat{f}$ according to (\ref{eqn:epe}) is that our initial model must be at least \emph{somewhat} informative to begin with. This typically entails choosing a small number of initial experiments to seed an initial starting model. Absent an informative model, we choose this small set of initial experiments to have \emph{low-discrepancy} over $\mathcal{D}$. This means for a budget of $N_{init}$ initial experiments, we aim to choose points $X_{init}$ such that in every subspace $\Delta D \subset \mathcal{D}$ the number of points in $X_{init}$ belonging to $\Delta D$ is proportional to the size of $\Delta D$ \cite{garud2017design}.

Standard methods exist to generate low discrepancy unit hyper-cube designs. For example, the \emph{Good Lattice Point Method} (\textsc{glp}) \cite{zaremba1966good} leverages the relative primes of $N_{init}$ to produce deterministic low-discrepancy designs. 

Robot environment domains are not typically perfect unit hypercubes---physical and spatial properties often have non-unit range limits, or constraints which cut across dimensions. Fortunately, we have shown in previous work \cite{innes2021automatic}, that one can transform a uniform design on the unit hypercube $\mathcal{U}_{init} = \{ u^{(1)} \dots u^{((N_{init})}\}$ into one for an arbitrary robotic domain using inverse rosenblatt transformation \cite{zhang2020construction}. As long as we are able to produce a cumulative distribution function (\textsc{cdf}) $F(x) = F_{1}(x_1)F_{2|1}(x_2 \mid x_1)F_{d|1\dots d-1}(x_d \mid x_1 \dots x_{d-1})$ over our desired space, we can transform one into the other as follows:

\begin{equation}
\begin{cases}
    \quad x^{(i)}_1 = F_{1}^{-1}(u^{(i)}_1) \\
    \quad x^{(i)}_j = F_{j|1 \dots j-1}^{-1}(u_j^{(i)} \mid x^{(i)}_1 \dots x^{(i)}_{j-1}),\  j = 2,\dots,d \\
\end{cases}
\label{eqn:inv-cdf}
\end{equation}

Where $F^{-1}(u)$ denotes the inverse of \textsc{cdf} $F(x)$.

\subsection{The full algorithm}
\label{sec:full-alg}

Algorithm (\ref{alg:full-system}) summarizes the process of generating adaptive experiment designs. Given an \textsc{stl} specification $\varphi$, a large set of candidate inputs $\textit{cand-exps}$, an expensive simulation $g$, initial experiments number $N_{init}$ and budget $N$, the algorithm proceeds as follows: First, it generates a surrogate model from our initial uniform design of $X_{init}$. Then, while it has remaining budget, it calculates the balance factor $\alpha$ and CV errors $e_{CV}^2$ required to choose an ideal point according to (\ref{eqn:epe}). It then runs this chosen point through $g$, calculates the \textsc{agm} robustness score from trace $\tau$, then updates $\hat{f}$ according to this new $x$, $y$ pair. 

\begin{algorithm}
        \caption{Adaptive Experiment Design for \textsc{STL} Models}
        \label{alg:full-system}
        \begin{algorithmic}[1]
        \Function{adapt-exp}{$\varphi,\textit{cand-exps},N, N_{init},g$}
            \State $U_{init} \gets$ \Call{glp}{$N_{init}$}
            \State $X_{init} \gets$ \Call{inv-rosen}{$U_{init}$}
            \State $X \gets X_{init}$
            \State $\bm{y} \gets $ Robustness scores for $X_{init}$
            \State $\hat{f} \gets$ Eqn (\ref{eqn:fit-gp})
            \For{$i = 1$ to $N$}
                \State $\alpha \gets$ Eqn (\ref{eqn:mepe-alpha})
                \State $CV \gets$ Eqn (\ref{eqn:ecv})
                \State $x^{(i)} \gets \underset{x \in \textit{cand-exps}}{\mathrm{argmax}} \textsc{epe}^{\alpha}_{CV}(x)$ Eqn (\ref{eqn:epe})
                \State $y^{(i)} \gets \Ls(\varphi, g(x^{(i)}), 0)$
                \State $\langle X, \bm{y}_{true} \rangle \gets \langle X \cup x^{(i)}, \bm{y} \cup y^{(i)} \rangle$
                \State $\hat{f} \gets$ Eqn (\ref{eqn:fit-gp})
            \EndFor
            \State \Return $\langle X, \bm{y}_{true}, \hat{f} \rangle$
        \EndFunction
        \end{algorithmic}
\end{algorithm}

\section{Experiments}
\label{sec:experiments}

\begin{figure}
    \centering
    \begin{subfigure}[b]{0.32\linewidth}
        \centering
        \includegraphics[width=\textwidth, height=\textwidth]{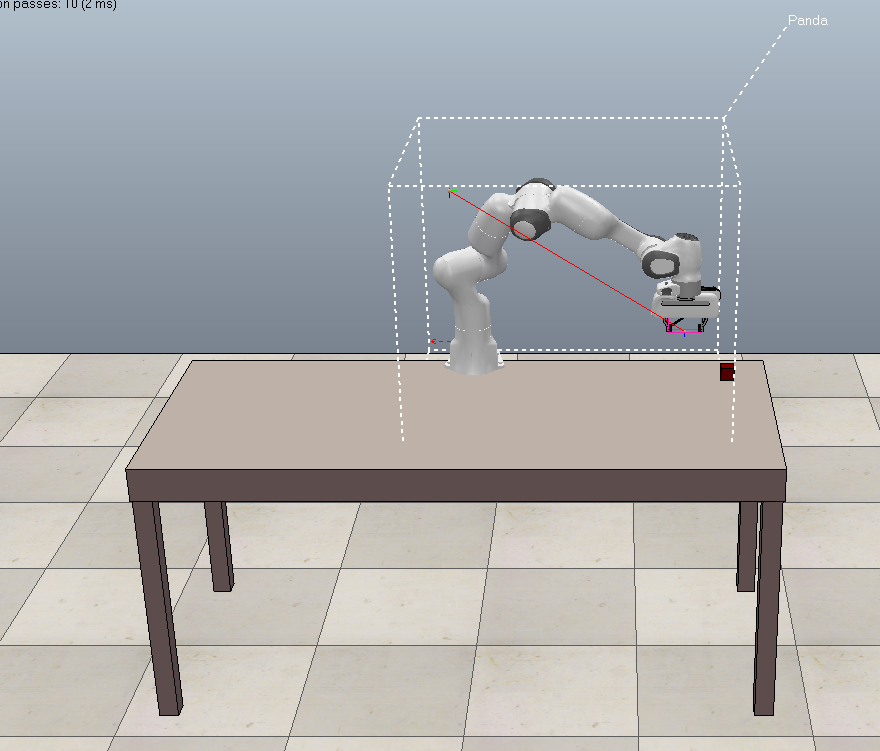}
        \caption{Reach}
        \label{fig:experiment-screenshots:reach}
    \end{subfigure}
    \begin{subfigure}[b]{0.32\linewidth}
        \centering
        \includegraphics[width=\textwidth, height=\textwidth]{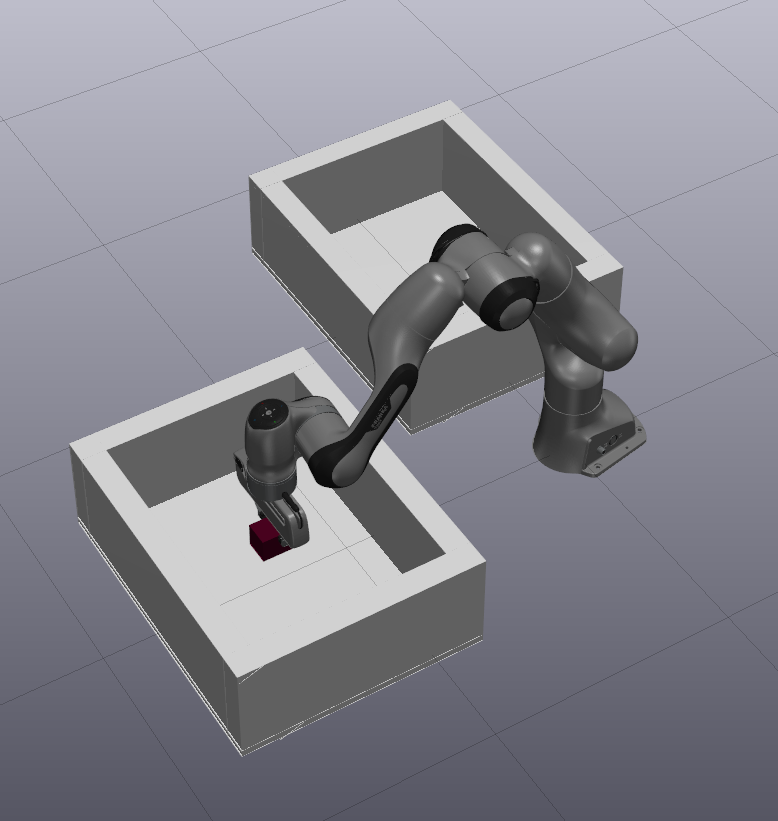}
        \caption{Pick and Place}
        \label{fig:experiment-screenshots:pick}
    \end{subfigure}
    \begin{subfigure}[b]{0.32\linewidth}
        \centering
        \includegraphics[width=\textwidth, height=\textwidth]{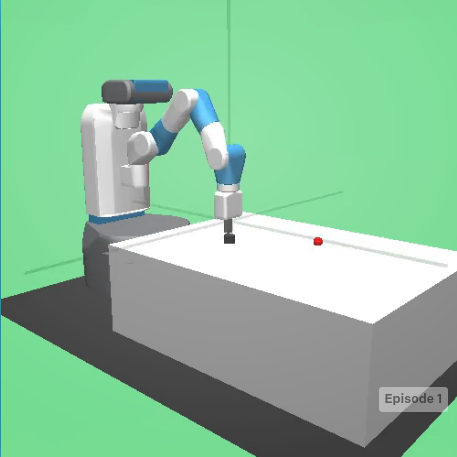}
        \caption{RL-Slide}
        \label{fig:experiment-screenshots:slide}
    \end{subfigure}
    \caption{Screenshots from the three experiment scenarios.}
    \label{fig:experiment-screenshots}
\end{figure}

\begin{figure*}
    \centering
    \begin{subfigure}[b]{0.32\linewidth}
        \centering
        \includegraphics[width=0.8\textwidth]{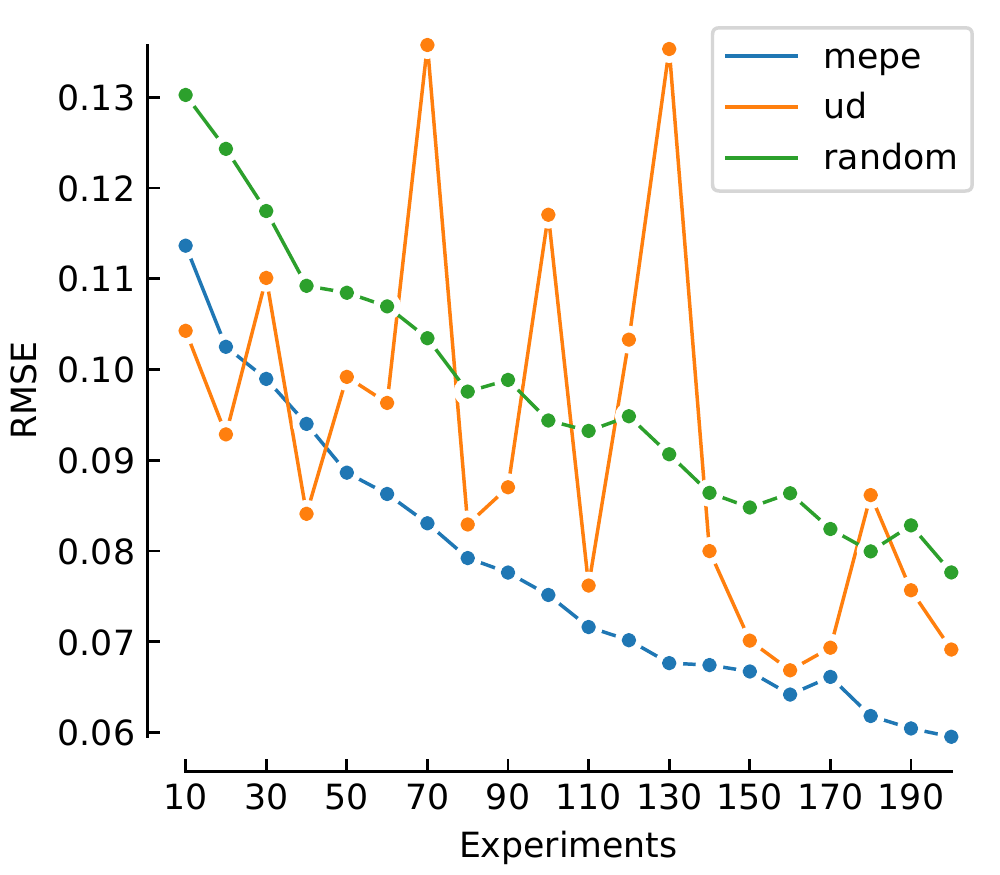}
        \caption{Reach}
        \label{fig:rmse-results:reach}
    \end{subfigure}
    \begin{subfigure}[b]{0.32\linewidth}
        \centering
        \includegraphics[width=0.8\textwidth]{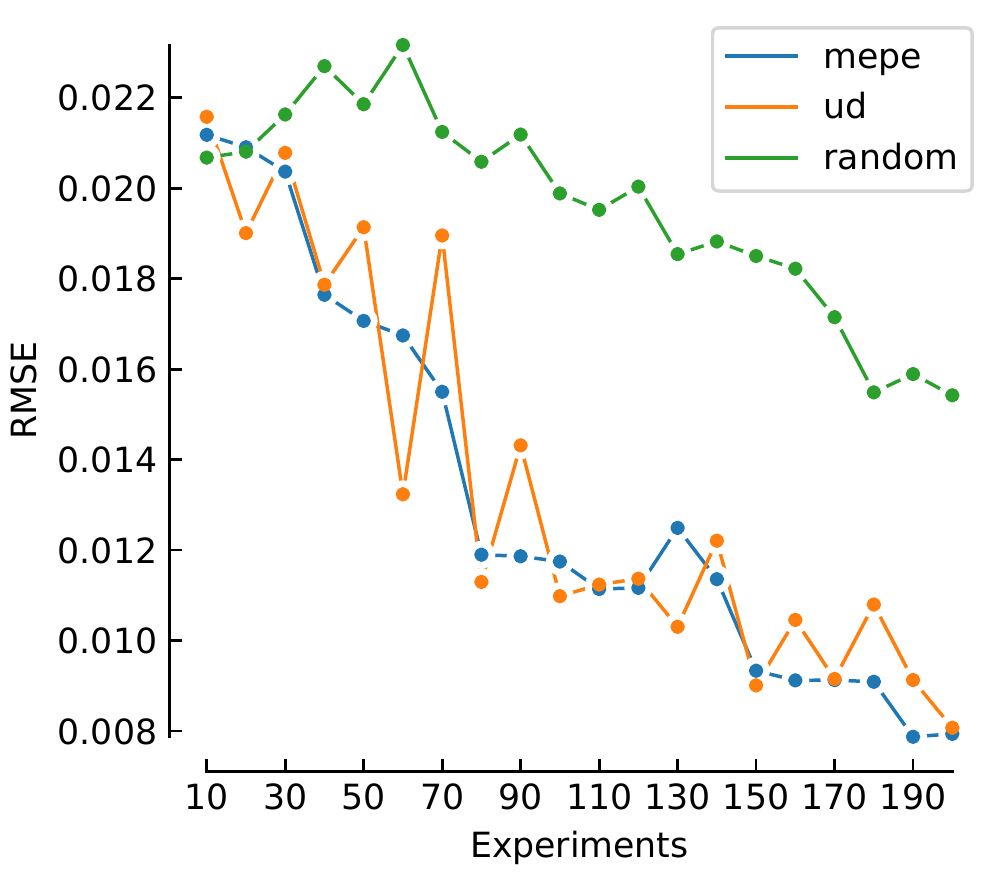}
        \caption{Pick-and-Place}
        \label{fig:rmse-results:pick}
    \end{subfigure}
    \begin{subfigure}[b]{0.32\linewidth}
        \centering
        \includegraphics[width=0.8\textwidth]{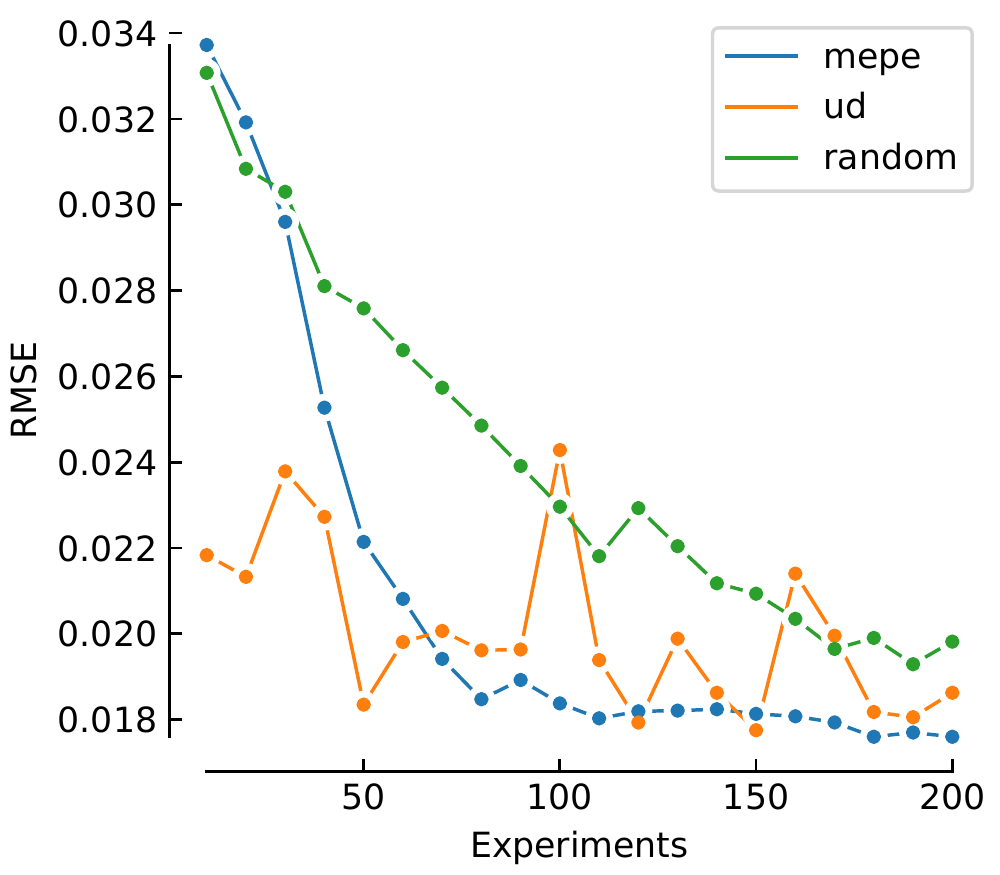}
        \caption{RL-Slide}
        \label{fig:rmse-results:slide}
    \end{subfigure}
    \caption{RMSE on $t=1000$ test points across the three experiment scenarios (\emph{random} results averaged over 30 repetitions).}
    \label{fig:rmse-results}
\end{figure*}

In this section, we evaluate algorithm (\ref{alg:full-system}) for three simulated robotic manipulation tasks --- reaching, pick-and-place, and sliding\footnote{Experiment code available at \url{https://github.com/craigiedon/automaticTestingSTL}}. The \textsc{stl} specifications $\varphi$ accompanying each task reflect common robot mission patterns (namely - \emph{Visitation}, \emph{Wait}, and \emph{Universal Liveness} \cite{menghi2019specification, dwyer1999patterns}). We show that, under various budgets $N$, our method chooses experiments that produce accurate surrogates for robustness across $\mathcal{D}$. Additionally, our method consistently outperforms both random sampling, and an $N$ experiment uniform design.

For each task, we generate $N_{init}=50$ initial experiments $\langle X_{init}, \bm{y}_{init} \rangle$ using the uniform design approach from section (\ref{sec:starting-exps}) to fit our initial $\hat{f}$. Here, $\hat{f}$ is a \textsc{gp} with a standard Matern kernel \cite{williams2006gaussian}, implemented using GPy\footnote{\url{https://sheffieldml.github.io/GPy/}}. We then compare the three methods for $N=[10, 20,  \dots 200]$. These methods are: \emph{random}, which samples experiments randomly from $\mathcal{D}$; \emph{ud}, which generates an $N$-experiment uniform design on $\mathcal{D}$ as described in (\ref{sec:starting-exps}); and \emph{mepe}, which maximizes the \textsc{epe} according to (\ref{eqn:epe}).

We evaluate method accuracy at $t=1000$ test points spread evenly across $\mathcal{D}$ by calculating average Root Mean-Squared Error (\textsc{rmse}) of the true robustness values $f(x)$ versus the ones estimated by $\hat{f}(x)$:

\begin{equation}
    \sqrt{\frac{1}{t} \sum_{i = 0}^{t} (f(x^{(i)}) - \hat{f}(x^{(i)})^2}
    \label{eqn:rmse}
\end{equation}

\subsection{Reach}
\label{sec:experiments:reach}

First, we consider a reaching task (Figure \ref{fig:experiment-screenshots:reach}). We simulate a 7-DOF Robotic Arm in CoppeliaSim \cite{rohmer2013coppeliasim}. The goal is to reach a target cube on the table. The associated task specification $\varphi$ is ``The gripper should be within $0.05$ metres of the target cube in at most 1 second'':

\begin{equation}
    \diamond_{[0, 1]} \left( \lVert r - c \rVert \leq 0.05 \right)
    \label{eqn:reach-spec}
\end{equation}

Here, $r$ is the $xyz$ coordinates of the robot's end-effector, and $c$ the $xyz$ coordinates of the cube. To solve the task, we use an \textsc{RRT} motion planner from the OMPL library \cite{sucan2012the-open-motion-planning-library}.

The 2-D domain $\mathcal{D}$ corresponds to the x/y starting positions of the cube on the surface of the table. We expect the true model $f(x)$ to essentially be a \emph{weighted reachability map} over table locations, where locations that take longer for the robot to reach have a smaller robustness, and unreachable areas have negative robustness.

\begin{figure}
    \centering
    \includegraphics[width=0.6\linewidth]{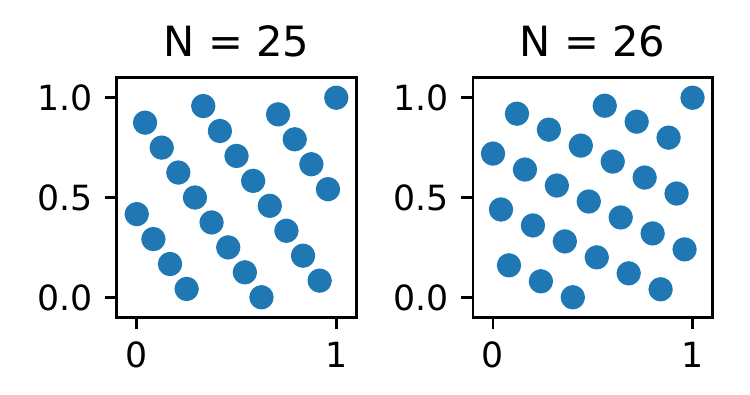}
    \caption{2-D uniform design staggerings for $N=\{25,26\}$.}
    \label{fig:staggerings}
\end{figure}

Figure (\ref{fig:rmse-results:reach}) shows the \textsc{rmse} of \emph{random}, \emph{ud}, and \emph{mepe} as the $N$ increases. The results show that, by leveraging model information, \emph{mepe} outperforms \emph{random} and \emph{ud} over the majority of budgets $N$. At low budgets (N=10-30), the difference between \emph{ud} and \emph{mepe} is negligible, as the model constructed by \emph{mepe} has too little data to assess which parts of $\mathcal{D}$ are interesting. However, as $N$ increases, $\hat{f}$ becomes more informative. This allows \emph{mepe} to select experiments in interesting areas $\mathcal{D}$, avoiding ``wasted'' experiments in uninteresting areas. A perhaps surprising aspect of these results is the relative stability of \emph{mepe} compared to \emph{ud}, which fluctuates as $N$ increases. As Figure (\ref{fig:staggerings}) illustrates, this is because while \emph{ud} reliably spreads its budget of experiments evenly across $\mathcal{D}$, the way in which points must be ``staggered'' to achieve even coverage can vary significantly for even small changes in $N$. This means the accuracy of models built using a uniform design often relies not just on the size of $N$, but on whether a given staggering happens to place points on interesting parts of $\mathcal{D}$.

\begin{figure}
    \centering
    \includegraphics[width=0.4\linewidth]{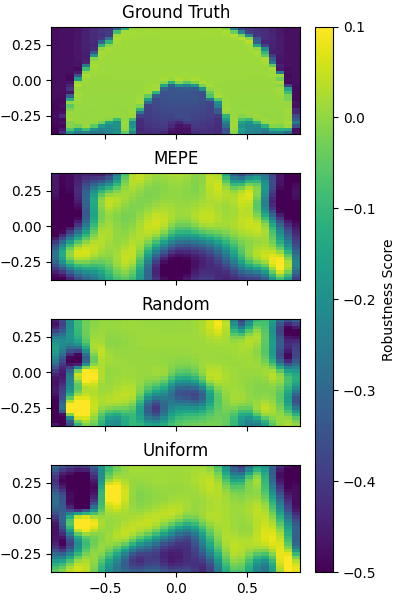}
    \caption{\emph{Reach} robustness as function of target xy ($N=50$).}
    \label{fig:reach-visualization}
\end{figure}

To understand how the models constructed by each approach look, Figure (\ref{fig:reach-visualization}) visualizes the predictions of each algorithm at $N = 50$. As expected, the ground-truth robustness for this task has positive robustness over most of the table, smaller values for points further away from the robot's start position, and negative robustness near the edges of the table where the cube is unreachable (meaning the robot cannot plan a path to that position). However, there is also a small dead-zone close to the robot's base, where the robot is unable to plan a path to the cube without self-collisions or joint limit violations. This results in an ``arc-like'' response for the robustness over $\varphi$. Comparing the models constructed by each method, we see that, even at this relatively low budget, \emph{mepe} has managed to capture this ``arc-like'' robustness landscape. By comparison, \emph{random} and \emph{ud} by have not yet captured this fundamental shape. 

\subsection{Pick-and-Place}
\label{sec:experiments:pick}

Our second experiment is a pick-and-place task in the Drake \cite{drake} simulator (Figure (\ref{fig:experiment-screenshots:pick})). The goal is to pick up a cube in one container, and move it to a target location in the other without dropping it. The associated $\varphi$ is ``If you grasp the cube, don't drop it until you reach the target'':

\begin{equation}
        \left( \mathit{grasp} \Rightarrow 
            \lVert r - c \rVert \leq \delta \right) \mathcal{U}_{[0, 30]} 
            \left( \lVert c - g  \rVert \leq \epsilon \right) 
        \label{eqn:pick-and-place-spec}
\end{equation}

Where $r$ is the xyz-position of the gripper, $c$ the xyz-position of the cube, and $g$ the goal location, with $\delta = 0.1$ and $\epsilon = 0.05$. The 1-D environment domain $\mathcal{D}$ consists of the mass $m \in [20, 70]$ in grams of the cube. We can now observe how our method performs with experiments on \emph{physical properties}, rather than spatial.

\begin{figure}
    \centering
    \includegraphics[width=0.4\linewidth]{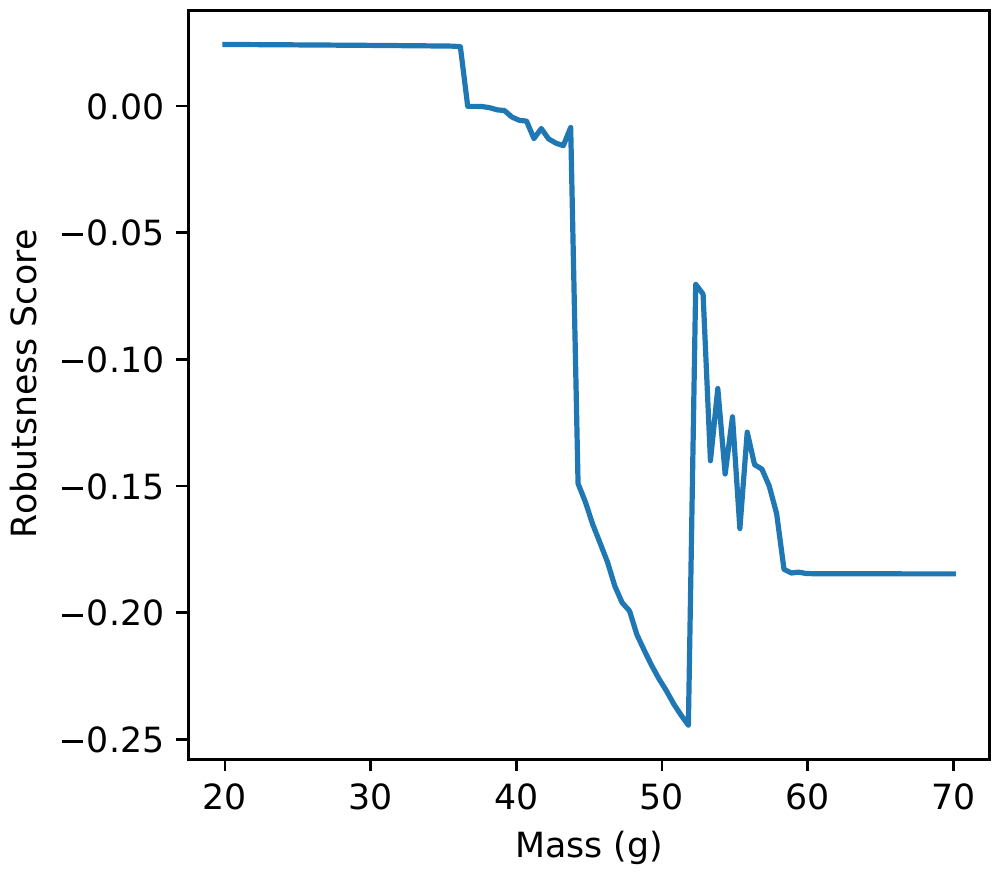}
    \caption{\emph{Pick-and-place} robustness for masses of 20-70g.}
    \label{fig:pick-mass-slice}
\end{figure}

One aim of this second experiment is to see if testing via specification model learning remains useful when there is a mismatch between the true $f$ and our model priors. To illustrate, Figure (\ref{fig:pick-mass-slice}) shows the true robustness our system as a function of $m$. For lighter masses ($20g-35g$), the robustness is flat---the controller easily completes the task. As the $m$ increases however, we encounter many sharp discontinuities, as the increasing mass of the cube causes it to slip from the robot's grasp at differing stages of its trajectory. At $m \geq 60$, the robustness again flattens---the cube is too heavy to lift. Such functions are challenging for our \textsc{gp}, which uses a \emph{stationary} kernel. In essence, this means the kernel assumes the ``wiggliness'' of $f$ is independent of the absolute value of $x$.

For this task, we use a pseudo-inverse based velocity controller for arm movement. For the gripper we use a velocity-based PD controller (which maintains a constant velocity without exceeding joint force limits). 

Figure (\ref{fig:rmse-results:pick}) gives the \textsc{rmse} scores. In contrast to the \emph{reach} experiment, \emph{mepe} and \emph{ud} now perform similarly, since our model requires many more samples before overcoming the mismatch between $f$ and the priors over $\hat{f}$. Despite this mismatch, both \emph{mepe} and \emph{ud} outperform \emph{random}, and use the budget to select experiments which give an informative view of robustness.

\subsection{RL-Slide}
\label{sec:experiments:slide}

Our third experiment demonstrates our method on ``black-box'' control algorithms learned via reinforcement learning (\textsc{rl}). We use the \emph{Fetch-Slide} environment from OpenAIGym's robotic task suite \cite{brockman2016openai}. The goal is to use the gripper to knock a puck to a target on the table. To achieve this, we use a publicly available pre-trained controller \footnote{\url{https://github.com/DLR-RM/rl-baselines3-zoo}} trained via Hindsight Experience Replay \cite{andrychowicz2017hindsight}. The associated $\varphi$ is ``Eventually, the puck should reach the target \emph{and stay there}'':

\begin{equation}
    \diamondsuit_{[0, 2]} \left( \square_{[0, 2]} \left( \lVert p - g \rVert \leq 0.1 \right) \right)
    \label{eqn:rl-slide-spec}
\end{equation}

Where $p$ and $g$ are the position of the puck and goal. For this experiment however, we wished to see how our method performed at assessing the robustness of the \textsc{rl}-controller \emph{not} on the \emph{original} environment, but on a \emph{modified} environment. Effectively, we emulate the effect of \emph{domain shift}. We modify the sliding friction parameter of the table in the original environment from $0.1$ to $0.06$, creating a ``slippery'' environment. The 2-D environment domain $\mathcal{D}$ is the xy-position of the goal. The start position of the puck is fixed. 

\begin{figure}
    \centering
    \includegraphics[width=0.4\linewidth]{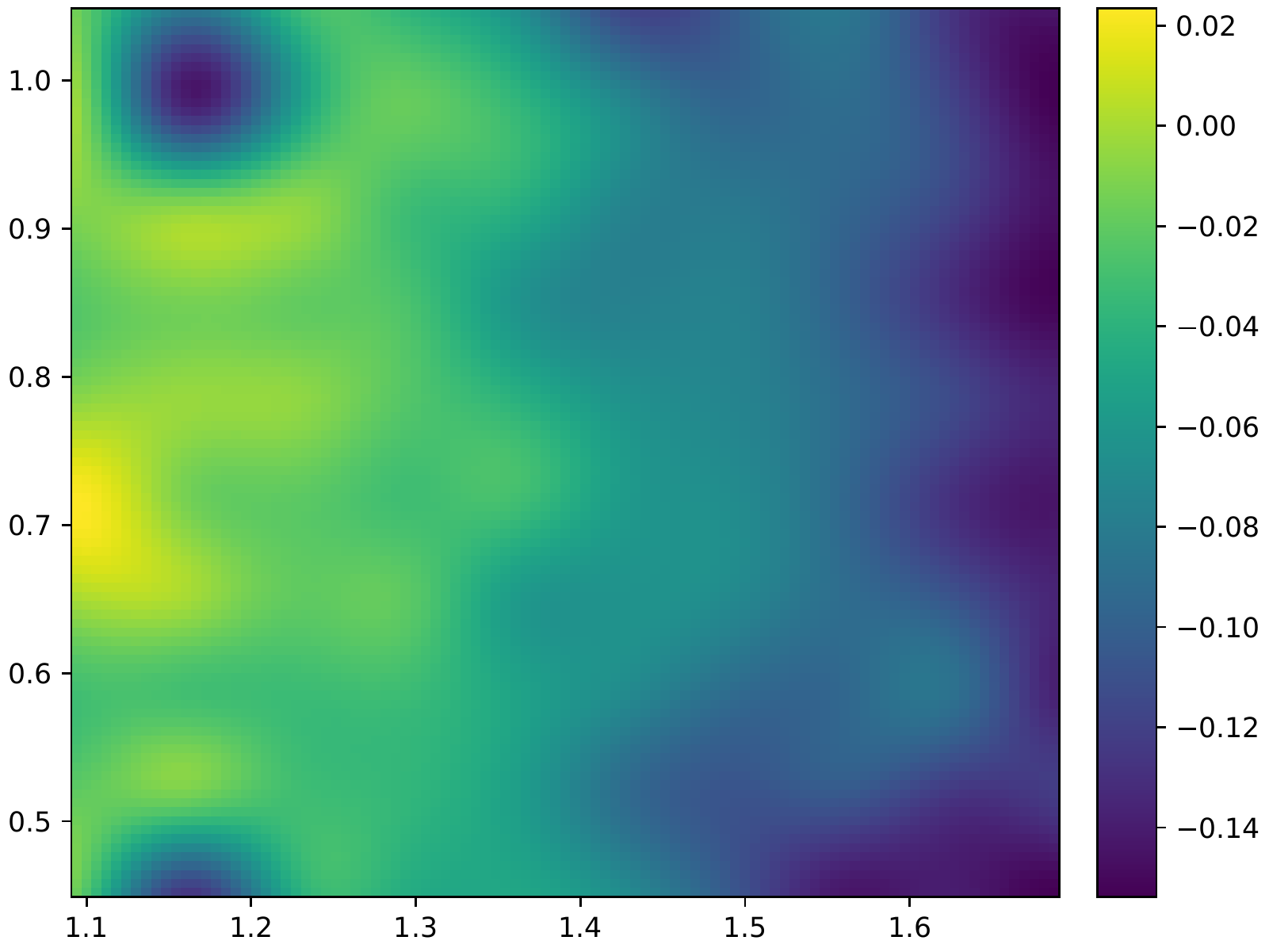}
    \caption{\emph{RL-Slide} \emph{mepe} robustness predictions ($N=60)$.}
    \label{fig:slide-viz}
\end{figure}

Figure (\ref{fig:rmse-results:slide}) shows the \textsc{rmse} results. Both \emph{mepe} and \emph{ud} consistently outperform \emph{random}. It takes \emph{mepe} 50-60 experiments to match (then outperform) \emph{ud}. More important than the individual results for this task is how we can \emph{use} $\hat{f}$ to assess controller robustness under domain-shift.

Figure (\ref{fig:slide-viz}) visualizes $\hat{f}$'s predictions as a function of goal position. Unsurprisingly, the model predicts negative values for goals near the right of the table (the robot arm is on the left side of the table)---the reduced friction causes our controller to overshoot, knocking the puck off the table. When the goal is on the left of the table, our controller corrects for overshooting by nudging the puck post-hit, allowing it to come to rest permanently at the goal. However, we discovered a somewhat surprising negative robustness ``hole'' in the top left corner. If the goal is nearly aligned with the x-position of the puck, but far away in the y-position, the robot aggressively overshoots, and is unable to move quickly behind the puck to correct itself. Such unexpected insights show one advantage of our testing method. It may provide useful information for future targeted re-training of our \textsc{rl}-controller in domain-shifted environments.

A natural question is whether it makes sense to impose a budget of $N$ for simulated \textsc{rl} scenarios. After all---the training algorithm likely ran thousands of simulations to optimize the controller's \textsc{rl} policy. While this is true for the example presented here, a more common setup is to first train an \textsc{rl}-algorithm on a fast, cheap simulator, then test it on a real robot. In such scenarios, each experiment becomes costly. Even resetting the environment between experiments requires costly manual labor or complex mechanisms \cite{zeng2020tossingbot}.

\section{Related Literature}
\label{sec:related-lit}

There are several techniques to falsify properties of ``black box'' machine learning (\textsc{ml}) systems \cite{dreossi2019compositional, dreossi2018semantic, annpureddy2011s, sankaranarayanan2012falsification, tuncali2018simulation}. These ``adversarial'' methods back-propagate robustness gradients through the \textsc{ml} components to find counterexamples which minimize robustness. However, unlike our method, they do not give an overall picture of robustness across $\mathcal{D}$. More generally, \emph{Bayesian optimization} \cite{greenhill2020bayesian} builds models to efficiently find global minima in high-dimensional space; \emph{Design space exploration} methods learn a \emph{pareto frontier} \cite{nardi2019practical, nardi2019hypermapper}---the perimeter along which multiple objectives are optimal. The core principle of balancing local and global search is similar to this paper, but models produced via these techniques are not useful summaries of the entire system---they are specifically tailored for \emph{optimization}. 

Given traces from a system with unknown dynamics, sensitivity analysis methods provide probabilistic bounds on how outputs change given small changes in input \cite{fan2016locally, fan2017d}. Such analyses could be used to verify whether an \textsc{stl}-specification will hold on a subset of inputs, and describe how likely this behaviour is to generalize within certain bounds. Such work complements ours well--- our method helps choose representative inputs to achieve good coverage over $\mathcal{D}$; while theirs takes those fixed inputs and makes descriptive statements about their probabilistic bounds.

The \textsc{mepe} metric used in this paper chooses experiments based on variance estimates from a \textsc{gp}, but other methods exist which choose experiments based on geometric features like locally-linear approximations of the output, or taylor-expansion based descriptors of nonlinearity \cite{crombecq2011novel, mo2017taylor}. Such methods often rely on having reliable gradient estimates, and are typically slower and less accurate than their variance-based counterparts \cite{fuhg2021state}. However, evaluating the utility of such techniques for robot control domains could make for interesting future work.

Our experiments used \textsc{gp}s to model specification robustness. An alternative would be use \emph{Neural Networks} (\textsc{NN}s). While \textsc{gp}s provide uncertainty estimates as a by-product of model construction, \textsc{NN}s typically do not. However, $s^2(x)$ can often be approximated by \emph{ensembling}. Here, multiple $\textsc{NN}s$ are trained either with differing hyper parameters, or data subsets. The inputs on which the \textsc{NN}s disagree are considered to have high uncertainty \cite{ipek2006efficiently}, \cite{eason2014adaptive}, \cite{jin2016adaptive}. $\textsc{NN}s$ tend to be ``data-hungry'', requiring thousands of data points to train. This makes them less viable for limited budget testing as considered in this paper.

Given limited budget, this paper's aim was to choose informative experiments. An alternative aim would be to reduce the time taken per experiment by learning a model of the simulation. Recent work uses graph-based \textsc{NN}s to learn fast approximations of a variety of physical phenomena \cite{battaglia2016interaction, li2018learning, chang2016compositional, pfaff2020learning, sanchez2020learning}. Such an approach presents numerous challenges. For one, such models require enormous amounts of data from a high-fidelity simulator to train---something we explicitly try to avoid. Additionally it is unclear whether current models generalize well to differing physical properties or time scales to those seen in the data. Another approach is to learn a fast model of the simulation which is ``just good enough" to catch violations of the specification \cite{menghi2020approximation}, and which is only refined if it starts to report spurious violations.

To assess \textsc{stl} satisfaction, we used \textsc{agm}-robustness \cite{mehdipour2019arithmetic}. Other metrics exist, which make different trade-offs in terms of soundness, temporal/spatial robustness, and learnability \cite{varnai2020robustness, mehdipour2019arithmetic, cohen2021model, donze2010robust, lindemann2019robust, pant2017smooth, haghighi2019control}. Such metrics are traditionally employed exclusively for falsification or controller optimization. It would be interesting to compare the empirical performance of such metrics for our task of learning a model of robustness across $\mathcal{D}$.

A related problem to ours is finding an \textsc{stl} expression to explain system abnormalities \cite{deng2021interpretable}. Such work solves a reverse problem to this paper --- Our work takes a fixed $\varphi$ and chooses $N$ experiments which analyse the behaviour of a fixed controller; theirs takes a fixed set of \emph{experiments}, and chooses a $\varphi$ which best explains them.

\section{Conclusion}
\label{sec:conclusion}

This paper integrated techniques from surrogate modelling, temporal logic, and adaptive experiment design to test robotic controllers on a limited budget. Our experiments show our method effectively leverages information about the task specification to choose informative inputs for building a model of system robustness. We consistently outperform random sampling and (depending on degree of model-mismatch) frequently surpass uniform designs. We also showed how to use our method to analyse \textsc{rl}-controllers, identifying areas of poor robustness under domain shift.

\bibliographystyle{IEEEtran}
\bibliography{references}
\end{document}